\documentclass[letterpaper, 10 pt, conference]{ieeeconf}  

\IEEEoverridecommandlockouts                              

\usepackage{hyperref}
\hypersetup{
    colorlinks=true,
    linkcolor=black,    
    urlcolor=blue,
    }
\usepackage{graphicx}
\usepackage{subcaption}
\usepackage{booktabs}
\usepackage{epsfig} 
\usepackage{amsmath} 
\usepackage{amssymb}  
\usepackage{mathrsfs}
\usepackage{cleveref}
\usepackage{color}
\usepackage{calc}
\usepackage{dsfont}
\usepackage{bm}
\usepackage{float}
\usepackage{url}
\usepackage{bm}
\usepackage[font=small,skip=2pt]{caption}
\usepackage{xspace}
\usepackage[dvipsnames]{xcolor}

\def \FigureAbbreviaition {Fig.}
\newcommand{\figref}[1]{\FigureAbbreviaition\ \ref{#1}}

\usepackage{amsfonts} 
\title{\LARGE \bf Manipulation via Force Distribution at Contact}

\author{Haegu Lee, Yitaek Kim, Casper Hewson Rask, and Christoffer Sloth 
\thanks{Authors are with the Maersk Mc-Kinney Moller Institute, University of Southern Denmark, Denmark {\tt\small \{haeg,yik,chsl\}@mmmi.sdu.dk  \tt\small \{crask21\}@student.sdu.dk.}}
}

\begin{document}
\maketitle
\thispagestyle{empty}
\pagestyle{empty}


\begin{abstract}
Efficient and robust trajectories play a crucial role in contact-rich manipulation, which demands accurate modeling of object-robot interactions. Many existing approaches rely on point contact models due to their computational efficiency. Simple contact models are computationally efficient but inherently limited for achieving human-like, contact-rich manipulation, as they fail to capture key frictional dynamics and torque generation observed in human manipulation. This study introduces a Force-Distributed Line Contact (FDLC) model in contact-rich manipulation and compares it against conventional point contact models. A bi-level optimization framework is constructed, in which the lower-level solves an optimization problem for contact force computation, and the upper-level optimization applies iLQR for trajectory optimization. Through this framework, the limitations of point contact are demonstrated, and the benefits of the FDLC in generating efficient and robust trajectories are established. The effectiveness of the proposed approach is validated by a box rotating task, demonstrating that FDLC enables trajectories generated via non-uniform force distributions along the contact line, while requiring lower control effort and less motion of the robot.
\end{abstract}

\section{Introduction}\label{sec:introduction}

Handling contact is crucial for solving contact-rich manipulation tasks, which are becoming increasingly important with rising demand for complex automation. Model-based methods have shown the capability to generate reliable motions for contact-rich manipulation \cite{shirai2023tactile, shirai2024robust, leeIROS2025}. This progress has largely been achieved by relaxing contact models, both in locomotion and manipulation tasks \cite{kim2022contact, pang2023global}, enabling the generation of feasible trajectories even with complex hardware such as dexterous robotic hands \cite{suh2025dexterous}. Recent advances in Reinforcement Learning (RL) have enabled demonstration of complex dexterous manipulation, such as pen spinning \cite{ma2024eureka} and cube rotation \cite{OpenAI2019}.

The importance of generating high-quality trajectories is further emphasized as they can be used as warm start for RL \cite{yang2025physics} or as reference trajectories for model-based control method \cite{le2024fast}. Although relaxation of contact models enabled a shift from discrete methods such as Mixed-Integer Quadratic Programming (MIQP) to gradient-based methods such as Model Predictive Control (MPC) and Iterative Linear Quadratic Regulator (iLQR), many approaches treat contact as a point contact. While point contact models may be sufficient for legged robots with point feet \cite{10802542}, it fails to provide a physically accurate representation for tasks that require precise manipulation when the robot establishes more than a point contact.
Moreover, contact-rich manipulation often relies on tactile sensors, which inherently exhibit compliance as they are frequently fabricated from soft materials such as gels  \cite{yuan2017gelsight, alspach2019soft}. This means that, in practice, the contact between a robot and its environments is more likely to be line or patch contact rather a single point contact. To achieve human-level dexterity in manipulation task, it is essential to recognize the limitations of point contact model and explore alternative formulations.

 \begin{figure}[t]
    \centering
    \includegraphics[width=1\linewidth]{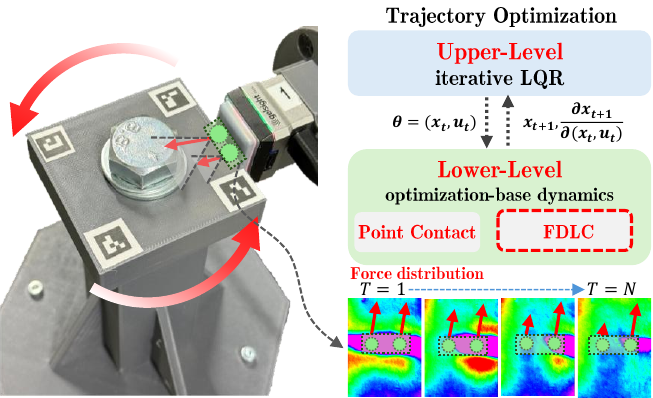}
    \vspace{0.05cm}
    \caption{Overview of the proposed framework combining iLQR, optimization-based dynamics with the Force-Distributed Line Contact (FDLC) model. Changes in force distribution at contact is shown}
    \label{fig:hero}
\end{figure}

Building on the idea of manipulation via force distribution at contact, we compare a conventional point contact model with a Force-Distributed Line Contact (FDLC) model. The FDLC is designed to mimic soft-surface interactions by distributing force along a contact line rather than concentrating it at a single point. To evaluate both contact models, we utilize optimization-based dynamics and iLQR with in a bi-level optimization framework from \cite{howell2022trajectory} to generate trajectories. Metrics, such as control effort, continuous force, goal tracking error and travel distance are introduced to compare the resulting trajectories. We assume that the surface of manipulated objects are planar. The overall framework and evaluation setup are shown in \figref{fig:hero}

\subsection{Contributions}
The main contributions of this work are summarized as follows:
\begin{itemize}
    \item \textbf{FDLC formulation}: We demonstrate the Force-Distributed Line Contact (FDLC) model for contact-rich manipulation, achieving lower control effort and indicating improved robustness to slip and geometry error in trajectory generation through iLQR optimization compared to point contact.
    \item \textbf{Simulation validation}: Demonstration of the enhanced torque application of the FDLC model through a comparative simulation analysis.
    \item \textbf{Experimental validation}: Proposed method was validated through experiments on real robotics setup, using trajectories generated by the framework.
\end{itemize}

This paper is organized as follows. Section~\ref{sec:related_works} presents related work to our research and Section~\ref{sec:method} introduces the
components and overall structure of the proposed method. In Section~\ref{sec:simulation} and \ref{sec:experimental_results}, the proposed method is verified with a simulation and real robotic setup. Finally, Section~\ref{sec:conclusions} concludes the paper and discusses future work.


\section{Related Work}\label{sec:related_works}
 Contact-rich manipulation requires quality reference trajectories to feed into offline controllers. Several studies have focused on getting trajectories that are specifically suitable for contact-rich manipulation \cite{manchester2019contact, shirai2025hierarchical, sleiman2019contact}. In many cases, interaction between the robot and objects is often modeled as point contact because of its computational efficiency. Such models are widely used in locomotion and simple non-prehensile tasks where torque generation can be neglected \cite{moura2022non, winkler2017fast}. In other cases, line contact is employed during the trajectory optimization to account for the geometry of the robot, but it is simplified into evenly distributed points with identical forces \cite{shirai2024robust}. 
  
Recent studies have proposed more explicit ways to present the line and patch contact to overcome the limitation of point contact. In the context of dynamical simulation, the Equivalent Contact Point (ECP) method has been introduced to parametrize the distributed forces and moments over a contact patch by a single representative point \cite{xie2016rigid}. Extended from this work, \cite{dietz2024high} utilizes the ECP method to model patch contacts and embed non-differentiable signed distance functions (SDF) into a complementarity Lagrangian System (CLS) to directly provide solutions to non-smooth rigid-body dynamics within optimal control problems. However, neither method has yet been extended to trajectory optimization problems with contact. The Pressure Field Contact (PFC) was proposed to incorporate contact with deformation and thereby predict contact surface, pressure distribution and net contact wrench \cite{elandt2019pressure, masterjohn2022velocity}. Although contact-implicit trajectory optimization can be formulated using PFC and iLQR \cite{kurtz2022contact}, its computation time is prohibitively slow for practical use, primarily due to the expensive gradient evaluations of hydroelastic contact in contact-rich environments. 

Improvements in tactile sensors and other compliant interfaces makes line or patch contact a more realistic assumption than point contact from a practical point of view \cite{hogan2020tactile, oller2024tactile}. Despite these advances, there are only a few studies that address line and patch contact modeling within trajectory optimization in a simple yet efficient manner. Our method employs the FDLC model that captures force distribution along a contact line. This approach enables the generation of efficient trajectories while preserving the advantages of the line contact model.  

\section{Method}\label{sec:method}
In this section, we present our methodology for generating trajectories for contact-rich manipulation tasks within a bi-level optimization framework. We consider both point contact and the proposed Force-Distributed Line Contact (FDLC) model. To address the limitation of the point contact model in representing realistic force interactions, the FDLC approximates line contact with two points connected by a virtual spring and damper. This formulation allows the two points to exert unequal forces, thereby representing the force distribution along the line. As a result, the FDLC model can represent torque generation through force distribution alone, without shifting the contact location, thus enabling the generation of efficient trajectories for contact-rich manipulation. The bi-level optimization approach from \cite{howell2022trajectory} is modified to incorporate FDLC model. In the following, we describe the point contact and FDLC models and provide a comparative analysis of their performance in trajectory optimization. 

\subsection{Force-Distributed Line Contact Model}\label{sec:contact_model}
The point contact model represents the interaction between the robot and the object as a single point of contact governed by friction cone constraints. Unless explicitly modeled as a soft finger, point contacts can exert forces but cannot directly produce contact torque, limiting their ability to achieve efficient rotational motion. 
 \begin{figure}[h]
    \centering
    \includegraphics[width=1\linewidth]{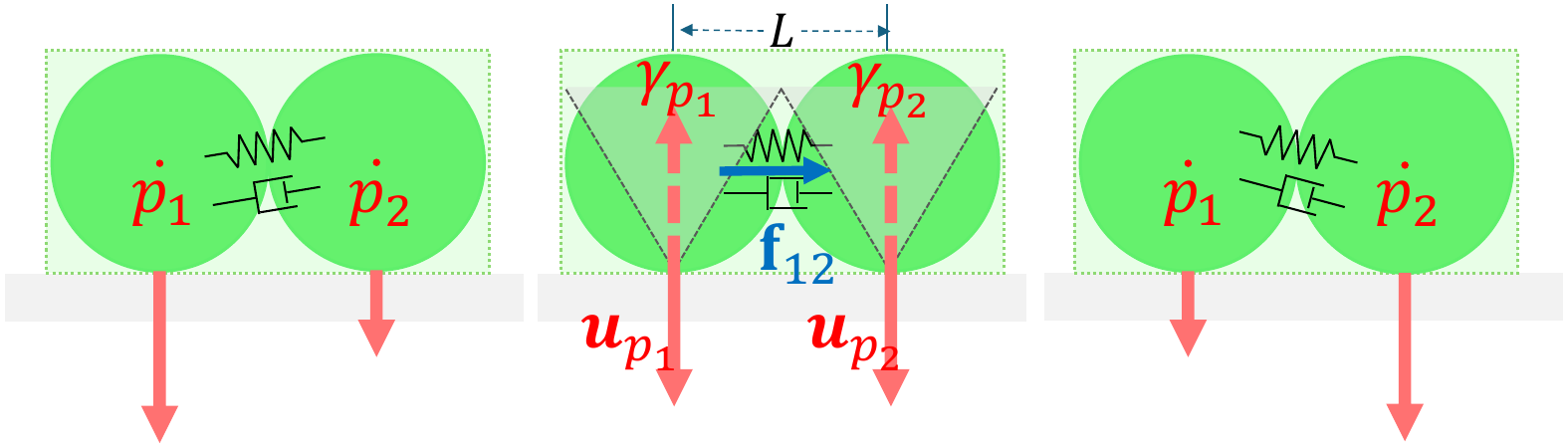}
    \caption{Illustration of the Force-Distributed Line Contact (FDLC) model, approximated by two points connected by a virtual spring–damper system, allowing for non-uniform force distribution.}
    \label{fig:line_contact}
\end{figure}

In contrast, as illustrated in \figref{fig:line_contact}, the FDLC approximates the line contact segment with two point contacts connected by a virtual spring-damper system to represent a nonuniform force distribution through the exerted forces. In this way, the two points can exert independent forces, capturing the compliance of the object or robot through the spring-damper connection, while maintaining the alignment as a line. 

The two points, $\mathbf{p}_1$ and $\mathbf{p}_2$, are connected by a virtual spring-damper element with stiffness $k$, damping $c$, and rest length $L$, to model the interaction forces $\mathbf{f}_{12}$ and $\mathbf{f}_{21}$:
\begin{align}
\mathbf{r} &= \mathbf{p}_2 - \mathbf{p}_1, 
\quad d = \|\mathbf{r}\|, 
\quad \mathbf{n} = \frac{\mathbf{r}}{d}, \\[6pt]
\mathbf{f}_{12} &= -k (d - L)\,\mathbf{n} 
 - c \left((\dot{\mathbf{p}}_2 - \dot{\mathbf{p}}_1)^{\top} \mathbf{n}\right)\mathbf{n}, \\[6pt]
\mathbf{f}_{21} &= -\mathbf{f}_{12}.
\end{align}
This virtual spring-damper formulation serves as the basis for the FDLC model and is incorporated into the system dynamics for the lower-level problem within the bi-level optimization framework.
Both point contact and FDLC are fundamentally constrained by the friction between the robot and the object.
The contact forces are determined implicitly by the lower level optimization, where each contact force consists of normal and tangential components, expressed as \(\mathbf{F}_{i}=[f_{n}^i, \bm{f}_{t}^i]^\top\).
The friction cone condition is given by
\begin{gather} 
\| f_i^t \| \leq \mu_i f_i^n, \quad f_i^n \geq 0, \quad \forall i \in \mathcal{C}\\
\mathcal{C} = 
\begin{cases}
A, & \text{point contact}, \\
B, C & \text{FDLC},
\end{cases}
\end{gather}
where $\|\bm{f}_{t}^i\|$ denotes the 2-norm of the tangential force at each contact point $i$, and \(\mu_i\) is the coefficient of friction.

The contact mode for FDLC can become overly complicated when approximated by $N_p$ points. However, trajectory optimization implicitly handles these complex contact modes, allowing feasible mode transitions without explicit definitions. In this paper, we restrict our study to the case \(N_p=2\), which is sufficient to capture the effect of force distribution. In addition, sliding along the contact surface is likely to occur simultaneously for the points, as they are connected. 

\subsection{Bi-level Optimization for Trajectory generation}
We employ a bi-level optimization framework where an upper-level optimizer makes use of the solution and gradients obtained from a lower-level optimization problem. The lower-level enforces the system dynamics with the chosen contact model (point contact or FDLC), while the upper-level trajectory optimization is carried out using the iLQR algorithm, which defined as follows:


\begin{align}
\min_{u_0:{T-1}} \quad 
& x_T^\top Q x_T 
+ \sum_{t=1}^{T-1} \Big( x_t^\top Q x_t 
+ u_t^\top R u_t 
+ \tfrac{1}{2} w \, \phi(x_t)^2 \Big) \label{eq:upper_obj} \\ 
\text{s.t.}\quad
& x_{t+1} = f(x_t, u_t), \quad t=1,...,T-1, \label{eq:upper_dyn}
\end{align}%
where $x_t \in \mathbb{R}^n$ is the state and $u_t \in \mathbb{R}^m$ is the control input, $\phi$ denotes the signed distance function, and $t$ is the time index. 
The upper-level optimizer aims to minimize both the terminal cost and the stage cost, 
each defined with quadratic weight matrices $Q$ and $R$, respectively. The Signed Distance Function (SDF) $\phi(x_t)$ is additionally introduced 
to penalize the separation between the robot and the object, thereby encouraging the robot to rotate the object while maintaining sufficient contact forces. The parameter $w \ge 0$ is a coefficient that modulates the relative influence of the distance cost with respect to the other costs. The additional term $\phi(x_t)$ does not necessarily contribute to achieving the final goal but rather encourages the robot to remain close to the object throughout the motion, which is often required in contact-rich and in-hand manipulation tasks.

The discrete dynamics in \eqref{eq:upper_dyn} are defined implicitly by a lower-level optimization-based dynamics problem formulated as follows:
\begin{equation}
    x_{t+1} \in z^{\ast}(\theta_p) \;=\; \label{eq:lower_obd}
    \operatorname*{arg\,min}_{z \in \mathcal{K},\, c(z;\theta_p)=0} \; \ell(z;\theta_p)
\end{equation}
where $z \in \mathbb{R}^k$ is the decision variable contains the next state and variables that represent contact force, frictional forces and slacks. $\theta \in \mathbb{R}^m$ is problem data that contains states, objectives, control input, etc. The objective function is defined as $\ell : \mathbb{R}^k \times \mathbb{R}^m \to \mathbb{R}$. The equality constraints are represented by $c: \mathbb{R}^k \times \mathbb{R}^m$ and $\mathcal{K}$ represents cone constraints, such as positive-orthant and second-order-cone constraints.
During rollouts, we evaluate \eqref{eq:lower_obd} to obtain $x_{t+1}$; the Jacobians 
$A_t=\partial x_{t+1}/\partial x_t$ and $B_t=\partial x_{t+1}/\partial u_t$ are computed by differentiating the lower-level optimality conditions via the implicit-function theorem and are passed to iLQR’s backward pass. We refer to \cite{howell2022trajectory}, \cite{howell2022dojo} for a more detailed explanations.

In this bi-level optimization framework, the FDLC model is incorporated into the lower-level residual, ensuring that optimization solver accounts for the coupling effects introduced by FDLC approximation. This allows the generation of trajectories that reflects force distributions, providing a basis for the comparative analysis presented in the following section.

\section{Simulation}\label{sec:simulation}
 \begin{figure}[h]
    \centering
    \includegraphics[width=1\linewidth]{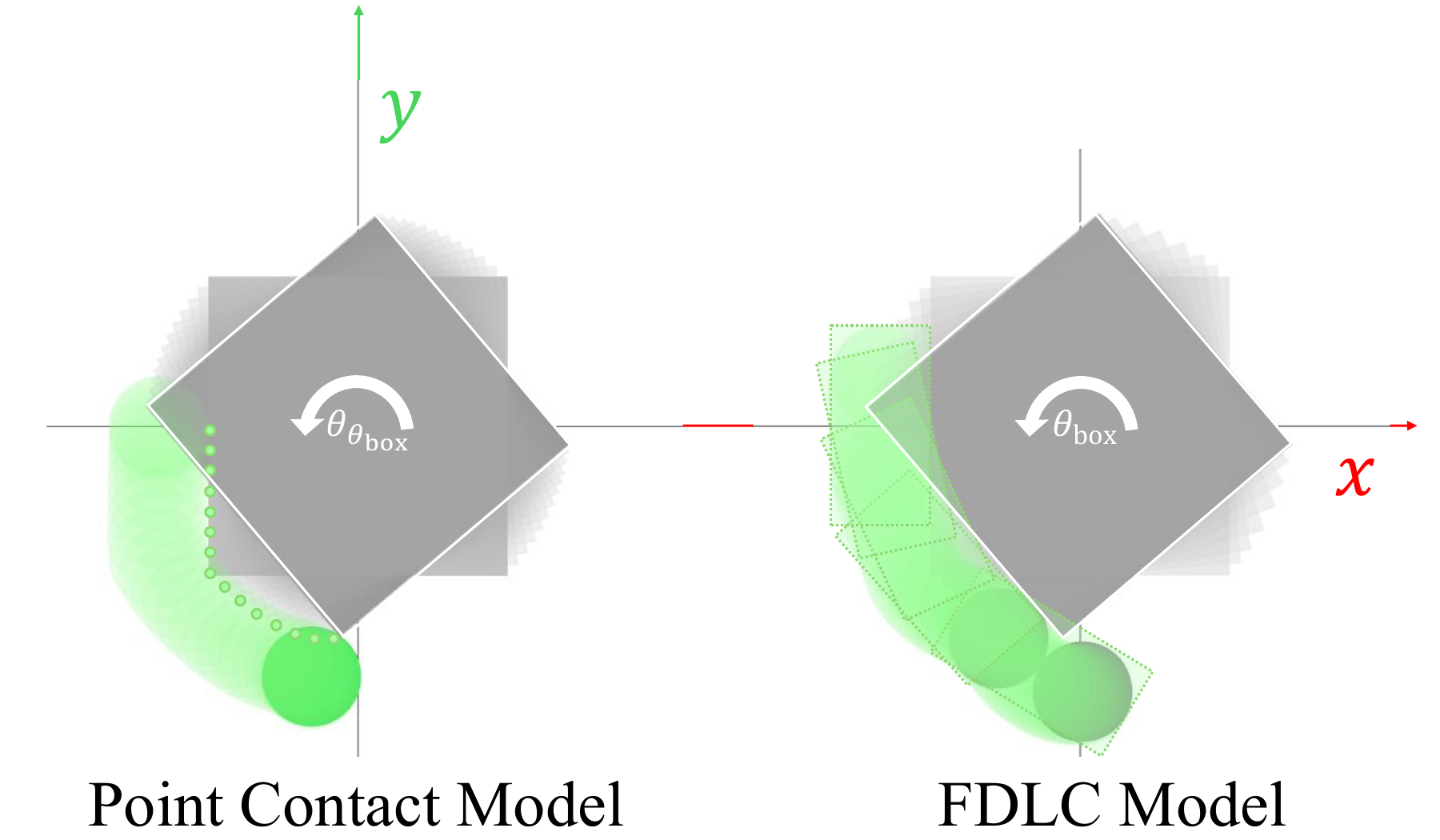}
    \vspace{-0.15cm}
    \caption{\small{Execution of trajectories generated by the point contact model (left) and the FDLC model (right) in the box rotation task.}}
    \label{fig:simulation}
\end{figure}
\begin{figure*}[t]
    \centering
    \includegraphics[width=1\linewidth]{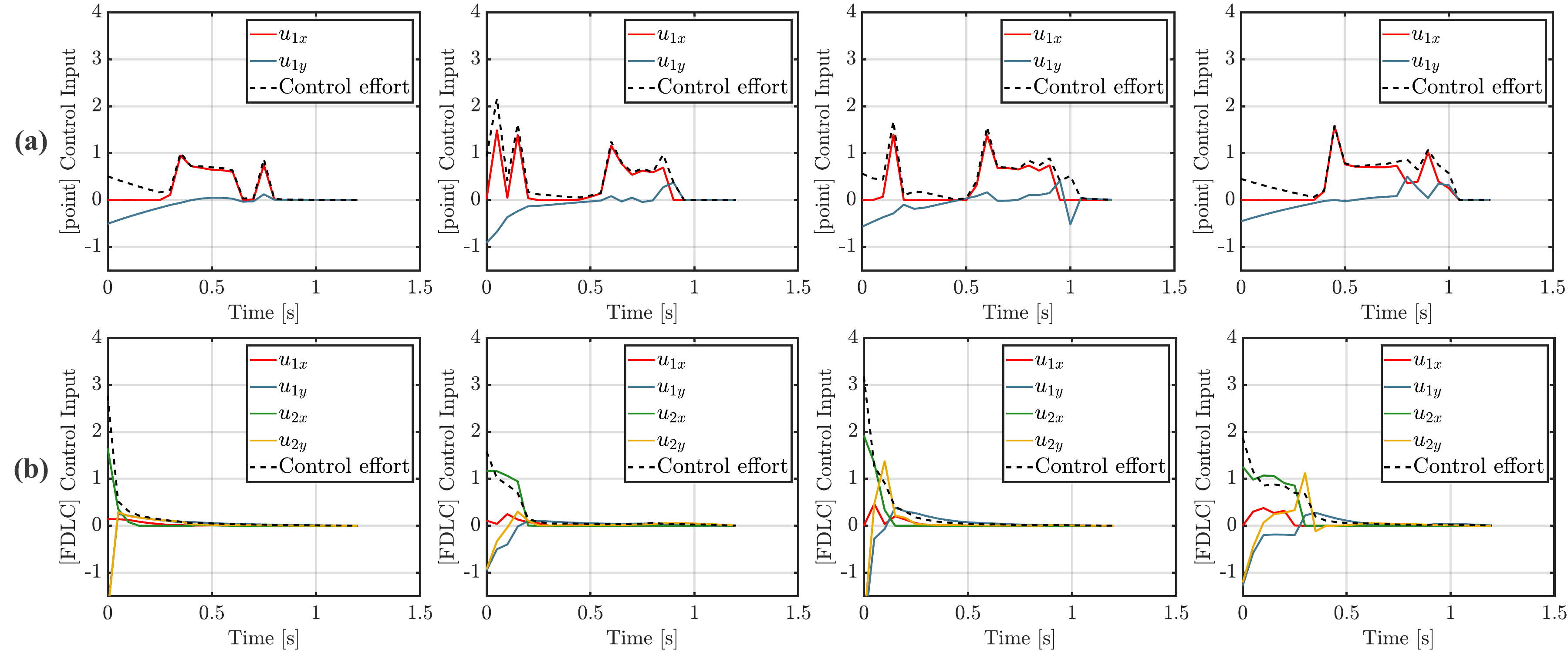}
    \caption{Control input from generated trajectory for each \(\theta_\textnormal{goal}\). The row (a) shows the control input with point contact model, while the row (b) shows the control input with FDLC model.}
    \label{fig:control_input}
    \vspace{-0.4cm}
\end{figure*}%
To evaluate the effectiveness of the proposed FDLC model for trajectory generation, the rotation of a box is simulated. The task environment is designed to highlight the differences between the point contact and FDLC models within a bi-level optimization framework. \figref{fig:simulation} illustrates the execution of trajectories generated with the point contact and the FDLC model. The box is constrained to the plane with its orientation defined along the \(z\)-axis. For simplicity, the translational states of the box, \(x_b\) and \(y_b\), are omitted, as the task requires only rotational state \(\theta_\textnormal{box}\). The robot is modeled as a circular virtual pusher with position [\(x_\textnormal{p1}, y_\textnormal{p1}\)]. 

Since the two points in FDLC can independently generate forces, the orientation of the robot is implicitly captured. While the virtual spring-damper connection might appear to imply a rotational degree of freedom, such orientation is not explicitly included in the model. 
Accordingly, the state and control input vectors for the optimization problem are defined as 
\begin{equation}
\begin{aligned}
\bm{x}_k &=
\begin{bmatrix}
\theta_\textnormal{box}, & x_{p_1}, & y_{p_1}, & \cdots & x_{p_{N_p}}, & y_{p_{N_p}}
\end{bmatrix}^\top, \\[6pt]
\bm{u}_k &=
\begin{bmatrix}
u_{x_1}, & u_{y_1}, & \cdots & u_{x_{N_p}}, & u_{y_{N_p}}
\end{bmatrix}^\top
\end{aligned}
\end{equation}%
where \(N_p = 1\) for the point contact model and \(N_p = 2\) for the FDLC model. Rotational friction of the box is represented by the translational friction force acting at its corners. To ensure a fair comparison, the same set of physical parameters were used to simulate both contact models. Detailed parameter values are provided in TABLE~\ref{table:parameters}.

The task is to rotate the box from its initial orientation of $\theta_\textnormal{box} = 0$ to target orientations of $\theta_\textnormal{goal} \in \{10^\circ, 20^\circ, 30^\circ, 40^\circ\}$ within a horizon of $T = 26$. The controller parameters are set with a time step $h = 0.05s$, using ${Q} = \mathrm{diag}(1,0.1,0.1)$ and ${R} = \mathrm{diag}(1,0.1,0.1)$ for the point contact model, and ${Q} = \mathrm{diag}(1,0.1,0.1,0.1,0.1)$ and ${R} = \mathrm{diag}(1,0.1,0.1,0.1,0.1)$ for the FDLC model. The initial control inputs are applied only along the $x$-axis with the same magnitude. We then conducted evaluations of both models.

\begin{table}[h]
\centering
\caption{Simulation system parameters.}
\begin{tabular}{l c c}
\hline
Property & Symbol & Value \\
\hline
Coefficient of friction (pusher--object) & $\mu_p$ & 0.5 \\
Coefficient of friction (object--surface) & $\mu_s$ & 1.0 \\
Mass of object, $kg$ & $m$ & 1 \\
Object side length, $m$ & $a$ & 0.02 \\
Line pusher width, $m$ & $d$ & 0.005 \\
\hline
\end{tabular}
\label{table:parameters}
\end{table}
\subsection{Control Effort and Travel Distance}
\begin{figure}[h]
    \centering
    \includegraphics[width=0.75\linewidth]{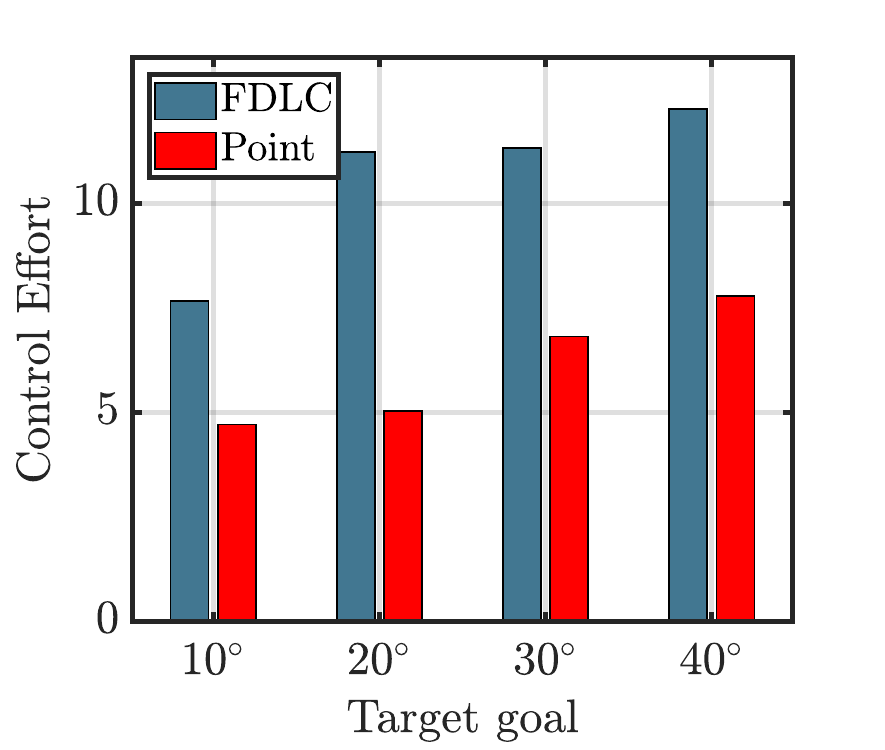}
    \caption{Control effort comparison between the FDLC model and the point contact model for different target angles.}
    \label{fig:control_effort}
    \vspace{-0.6cm}
\end{figure}
\begin{figure*}[t]
    \centering
    \includegraphics[width=1\linewidth]{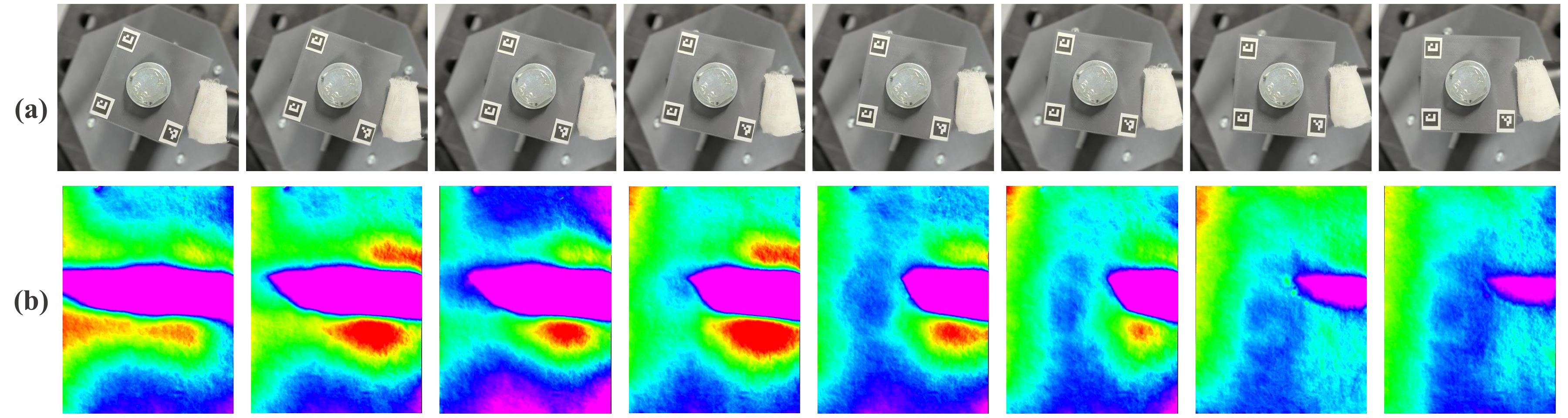}
    \caption{Control input from generated trajectory for each \(\theta_\textnormal{goal}\) The row (a) of the figure shows the control input with point contact model, while the row (b) shows the control input with line contact model.}
    \label{fig:actual_exp_with_force_dist}
    \vspace{-0.05cm}
\end{figure*}
The efficiency of the generated trajectory is evaluated by measuring the accumulated control effort over the time horizon, as well as travel distance \cite{zeng2017energy, kelly2017introduction}. Since the high-level optimization problem is formulated to minimize control effort, this serves as an ideal metric to compare the performance of different contact models. In contact-rich manipulation, a lower control effort indicates that a task can be achieved with less force. \figref{fig:control_input} shows the control input profiles for a trajectory corresponding to a target angle $\theta_d \in \{10^\circ, 20^\circ, 30^\circ, 40^\circ\}$. In \figref{fig:control_input}(b), trajectories generated with the FDLC model exhibit relatively high initial control input but rapidly decrease as the box begins to rotate. In contrast, as shown in \figref{fig:control_input}(a), the point contact model requires continuous inputs throughout the horizon due to the need for sliding along the contact surface to generate torque. This difference is shown in \figref{fig:control_effort}, which compares the accumulated control effort between the FDLC model and the point contact model. 

The underlying reason for this difference is that the FDLC can generate direct torque through force distribution along the contact segment. On the other hand, point contact relies on sliding to create an effective lever arm. Reduced control effort not only emphasizes the efficiency in simulation but also benefits for robotic systems. 

\begin{figure}[!h]
    \centering
    \includegraphics[width=1\linewidth]{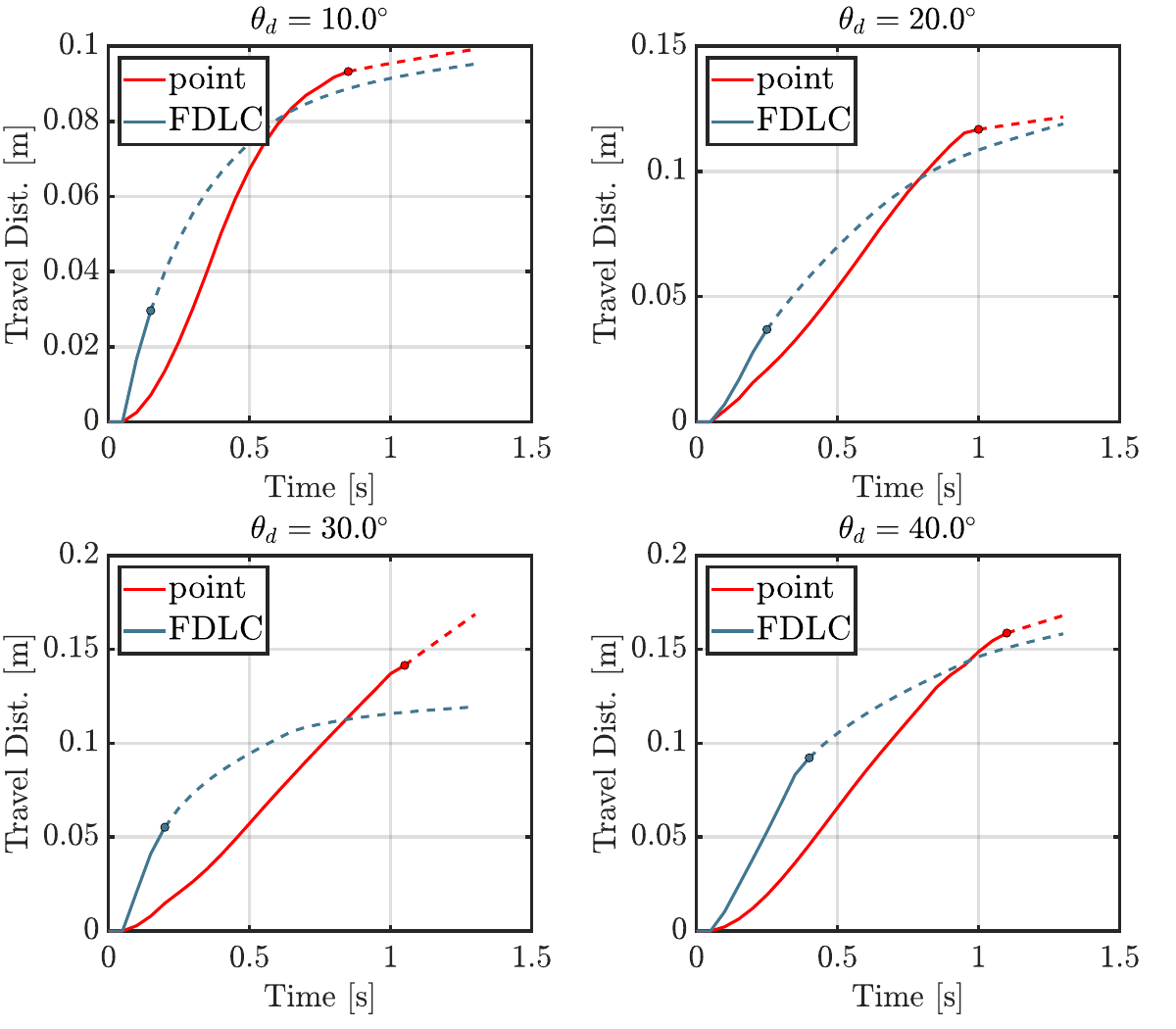}
    \caption{Travel distances of robot when executing trajectories generated using the point contact and FDLC model, with dots denoting the moment when the object reaches desired angle \(\theta_d\). }
    \label{fig:travel_distance}
\end{figure}

Additionally, \figref{fig:travel_distance} shows the travel distance of the robot when executing trajectories generated with the point contact and FDLC models for different desired angles. In all cases, the FDLC requires less travel distance than the point contact model to rotate the box. The dots in the figures indicate the moment when the box reaches the desired angle \(\theta_d\). This demonstrates that the FDLC can manipulate the object through force distribution rather than large end-effector movements.
\begin{figure}[h]
    \centering
    \includegraphics[width=1.05\linewidth]{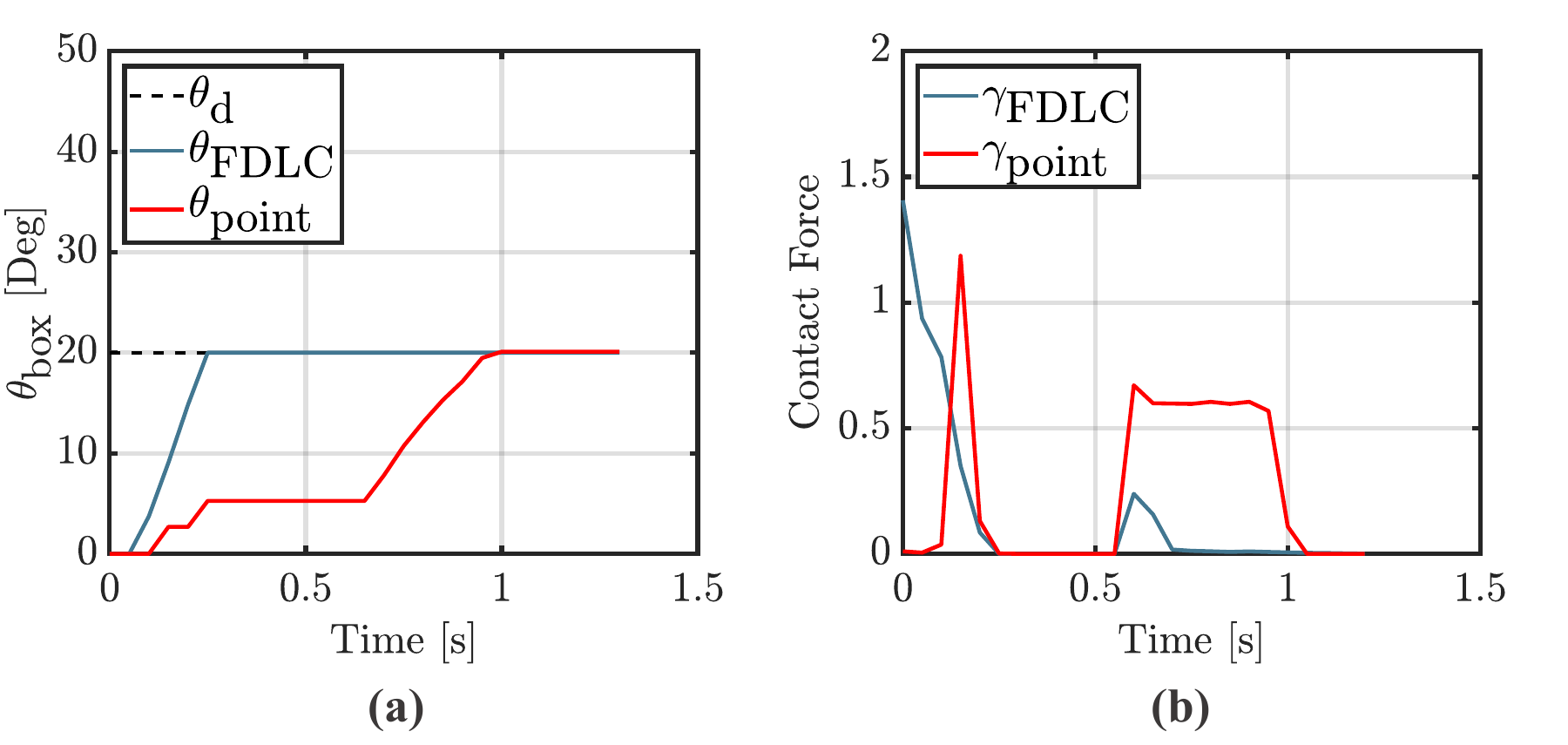}
    \caption{Desired box $\theta_d$ and simulation box orientation when executing trajectories generated from the FDLC model ($\theta_{\text{FDLC}}$) and point contact model ($\theta_{\text{point}}$). (b) Contact force generated from executing trajectories.}
    \label{fig:contact_force}
\end{figure}
\subsection{Contact Forces and Robustness}
Contact forces applied to the object during manipulation are a critical factor for robustness. \figref{fig:contact_force}(b) shows that the FDLC model generates continuous contact forces until the object angle \(\theta\) reaches the desired angle \(\theta_d\). This is possible because the FDLC model provides a slight geometric advantage by enabling force distribution, unlike the point contact model, which must slide along the contact surface to create a lever arm for torque application. 

The difference can be further explained from a force distribution perspective. As the FDLC model consists of two points that can generate unequal forces, force distribution that can maintain torque is well captured. By contrast, a point contact model suffers from contact mode changes, which result in intermittent force profiles.

In many contact-rich manipulation tasks, achieving high-precision control is challenging due to unavoidable slip and model uncertainty. As illustrated in \figref{fig:contact_force}, trajectories generated with the point contact model are more likely to have abrupt changes in the contact states, making them susceptible to unexpected slip. In contrast, the continuous force generation in the FDLC model reduces these effects, providing more robust motion against disturbances. From a practical standpoint, this robustness also makes the FDLC trajectories less sensitive to parameter errors and sensor noise, which in turn improves the likelihood of successful execution on real robotic systems. 
\section{Experimental Results}\label{sec:experimental_results}
This section presents the experiments conducted to evaluate the proposed method and discusses the results. First, we show the outcome of executing the trajectories generated with the point contact and FDLC models. We then analyze the force distribution by using a GelSight image to evaluate the approximated FDLC. 
\begin{figure}[!t]
    \centering
    \includegraphics[width=1\linewidth]{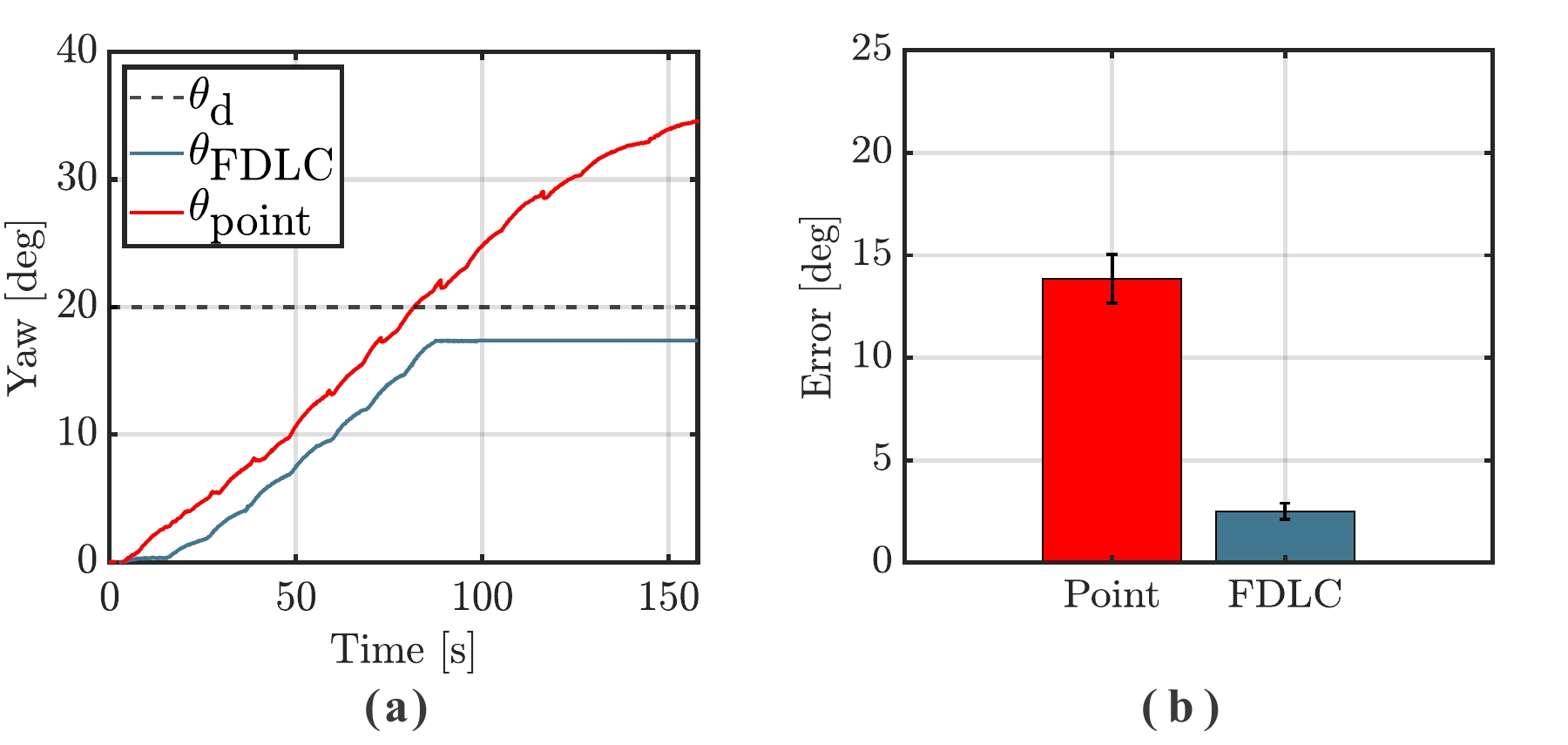}
    \caption{Desired box $\theta_d$ and measured yaw when executing trajectories generated from the FDLC model ($\theta_{\text{FDLC}}$) and point contact model ($\theta_{\text{point}}$). (b) Mean yaw error and standard deviation over five repetitions.}
    \label{fig:experiment_result}
\end{figure}
\subsection{Experiment Setup}
A Universal Robots UR5e equipped with a single GelSight sensor is used to observe the changes in the contact force distribution. We set up the experimental platform for validation, where the 3D-printed box is mounted on the table and can rotate under external forces as shown in Fig.~\ref{fig:hero}. In addition, we attach four ArUco markers on the box and install a single overhead camera that tracks them to provide the ground truth for the pose of the box. The bi-level trajectory optimization in \eqref{eq:upper_dyn} generates the reference trajectories with the point contact and FDLC models, respectively. We then send the reference trajectories to the robot in an open loop without sensor feedback, using sensor measurements only to evaluate the performance of the generated trajectories. An admittance controller is used to track the force reference trajectories. Lastly, we assess the performance of the reference trajectories generated by each contact model based on the error between the desired and current yaw angles.

\subsection{Open-Loop Controller and Force Distribution}
In this experiment, we show that the FDLC model represented by two points with a virtual spring-damper system produces more effective trajectories than the point contact model. 

\figref{fig:experiment_result} (a) shows that neither trajectory reached the desired angle \(\theta_\textnormal{d}\), unlike in simulation, due to sim-to-real gap. However, \figref{fig:experiment_result} (b) demonstrates that \(\theta_\textnormal{FDLC}\) is closer to \(\theta_\textnormal{d}\)  than \(\theta_\textnormal{point}\). The main reason for the poor performance of the point contact model is that the robot equipped with GelSight was oversimplified compared to its actual geometry. As a result, it could not follow the force reference trajectory of the point contact model. In contrast, the robot successfully followed the reference trajectory from the FDLC model by adjusting the force distribution at contact. 

The GelSight image during execution, shown in \figref{fig:actual_exp_with_force_dist} (b), indicates that the contact area naturally shifts from the center to the left. This shift reflects that the left approximated point exerted a larger force than right point, which generates the torque without sliding along the contact surface.
\section{Conclusions}\label{sec:conclusions}
This paper introduces a Force-Distributed Line Contact (FDLC) model for generating efficient and robust trajectories in contact-rich manipulation. The FDLC model is approximated by two points connected through a spring-damper system, capturing changes in force distribution during manipulation. In the considered rotation of a box task, the FDLC model produced more efficient and robust trajectories compared to the point contact model, as it could generate the required torque without switching contact modes. This resulted in lower control effort, shorter travel distance and continuous contact force generation, making the system more robust against slip during the execution. 

For future work, we plan to incorporate an online controller to validate the effectiveness of high quality reference trajectories generated with the FDLC model. 

\bibliographystyle{IEEEtran}
\bibliography{reference}

@article{suh2025dexterous,
  title={Dexterous contact-rich manipulation via the contact trust region},
  author={Suh, HJ and Pang, Tao and Zhao, Tong and Tedrake, Russ},
  journal={arXiv preprint arXiv:2505.02291},
  year={2025}
}

@article{pang2023global,
  title={Global planning for contact-rich manipulation via local smoothing of quasi-dynamic contact models},
  author={Pang, Tao and Suh, HJ Terry and Yang, Lujie and Tedrake, Russ},
  journal={IEEE Transactions on robotics},
  volume={39},
  number={6},
  pages={4691--4711},
  year={2023},
  publisher={IEEE}
}

@inproceedings{sleiman2019contact,
  title={Contact-implicit trajectory optimization for dynamic object manipulation},
  author={Sleiman, Jean-Pierre and Carius, Jan and Grandia, Ruben and Wermelinger, Martin and Hutter, Marco},
  booktitle={2019 IEEE/RSJ international conference on intelligent robots and systems (IROS)},
  pages={6814--6821},
  year={2019},
  organization={IEEE}
}

@article{manchester2019contact,
  title={Contact-implicit trajectory optimization using variational integrators},
  author={Manchester, Zachary and Doshi, Neel and Wood, Robert J and Kuindersma, Scott},
  journal={The International Journal of Robotics Research},
  volume={38},
  number={12-13},
  pages={1463--1476},
  year={2019},
  publisher={SAGE Publications Sage UK: London, England}
}

@inproceedings{elandt2019pressure,
  title={A pressure field model for fast, robust approximation of net contact force and moment between nominally rigid objects},
  author={Elandt, Ryan and Drumwright, Evan and Sherman, Michael and Ruina, Andy},
  booktitle={2019 IEEE/RSJ International Conference on Intelligent Robots and Systems (IROS)},
  pages={8238--8245},
  year={2019},
  organization={IEEE}
}

@article{shirai2025hierarchical,
  title={Hierarchical Contact-Rich Trajectory Optimization for Multi-Modal Manipulation using Tight Convex Relaxations},
  author={Shirai, Yuki and Raghunathan, Arvind and Jha, Devesh K},
  journal={arXiv preprint arXiv:2503.07963},
  year={2025}
}

@article{shirai2024robust,
  title={Robust pivoting manipulation using contact implicit bilevel optimization},
  author={Shirai, Yuki and Jha, Devesh K and Raghunathan, Arvind U},
  journal={IEEE Transactions on Robotics},
  volume={40},
  pages={3425--3444},
  year={2024},
  publisher={IEEE}
}

@article{masterjohn2022velocity,
  title={Velocity level approximation of pressure field contact patches},
  author={Masterjohn, Joseph and Guoy, Damrong and Shepherd, John and Castro, Alejandro},
  journal={IEEE Robotics and Automation Letters},
  volume={7},
  number={4},
  pages={11593--11600},
  year={2022},
  publisher={IEEE}
}

@inproceedings{kurtz2022contact,
  title={Contact-implicit trajectory optimization with hydroelastic contact and ilqr},
  author={Kurtz, Vince and Lin, Hai},
  booktitle={2022 IEEE/RSJ International Conference on Intelligent Robots and Systems (IROS)},
  pages={8829--8834},
  year={2022},
  organization={IEEE}
}

@article{winkler2017fast,
  title={Fast trajectory optimization for legged robots using vertex-based zmp constraints},
  author={Winkler, Alexander W and Farshidian, Farbod and Pardo, Diego and Neunert, Michael and Buchli, Jonas},
  journal={IEEE Robotics and Automation Letters},
  volume={2},
  number={4},
  pages={2201--2208},
  year={2017},
  publisher={IEEE}
}

@inproceedings{moura2022non,
  title={Non-prehensile planar manipulation via trajectory optimization with complementarity constraints},
  author={Moura, Joao and Stouraitis, Theodoros and Vijayakumar, Sethu},
  booktitle={2022 International Conference on Robotics and Automation (ICRA)},
  pages={970--976},
  year={2022},
  organization={IEEE}
}

@article{howell2022trajectory,
  title={Trajectory optimization with optimization-based dynamics},
  author={Howell, Taylor A and Le Cleac’h, Simon and Singh, Sumeet and Florence, Pete and Manchester, Zachary and Sindhwani, Vikas},
  journal={IEEE Robotics and Automation Letters},
  volume={7},
  number={3},
  pages={6750--6757},
  year={2022},
  publisher={IEEE}
}

@article{howell2022dojo,
  title={Dojo: A differentiable simulator for robotics},
  author={Howell, Taylor A and Le Cleac’h, Simon and Kolter, J Zico and Schwager, Mac and Manchester, Zachary},
  journal={arXiv preprint arXiv:2203.00806},
  volume={9},
  number={2},
  pages={4},
  year={2022}
}

@INPROCEEDINGS{10802542,
  author={Ghansah, Adrian B. and Kim, Jeeseop and Li, Kejun and Ames, Aaron D.},
  booktitle={2024 IEEE/RSJ International Conference on Intelligent Robots and Systems (IROS)}, 
  title={Dynamic Walking on Highly Underactuated Point Foot Humanoids: Closing the Loop between HZD and HLIP}, 
  year={2024},
  volume={},
  number={},
  pages={12686-12693},
  doi={10.1109/IROS58592.2024.10802542}}

@article{yuan2017gelsight,
  title={Gelsight: High-resolution robot tactile sensors for estimating geometry and force},
  author={Yuan, Wenzhen and Dong, Siyuan and Adelson, Edward H},
  journal={Sensors},
  volume={17},
  number={12},
  pages={2762},
  year={2017},
  publisher={MDPI}
}

@article{OpenAI2019,
author = {OpenAI: Marcin Andrychowicz and Bowen Baker and Maciek Chociej and Rafal Józefowicz and Bob McGrew and Jakub Pachocki and Arthur Petron and Matthias Plappert and Glenn Powell and Alex Ray and Jonas Schneider and Szymon Sidor and Josh Tobin and Peter Welinder and Lilian Weng and Wojciech Zaremba},
title ={Learning dexterous in-hand manipulation},

journal = {The International Journal of Robotics Research},
volume = {39},
number = {1},
pages = {3-20},
year = {2020},
doi = {10.1177/0278364919887447},

URL = { 
    
        https://doi.org/10.1177/0278364919887447
    
    

},
eprint = { 
    
        https://doi.org/10.1177/0278364919887447
    
    

}
}

@inproceedings{
ma2024eureka,
title={Eureka: Human-Level Reward Design via Coding Large Language Models},
author={Yecheng Jason Ma and William Liang and Guanzhi Wang and De-An Huang and Osbert Bastani and Dinesh Jayaraman and Yuke Zhu and Linxi Fan and Anima Anandkumar},
booktitle={The Twelfth International Conference on Learning Representations},
year={2024},
url={https://openreview.net/forum?id=IEduRUO55F}
}

@inproceedings{alspach2019soft,
  title={Soft-bubble: A highly compliant dense geometry tactile sensor for robot manipulation},
  author={Alspach, Alex and Hashimoto, Kunimatsu and Kuppuswamy, Naveen and Tedrake, Russ},
  booktitle={2019 2nd IEEE International Conference on Soft Robotics (RoboSoft)},
  pages={597--604},
  year={2019},
  organization={IEEE}
}

@article{le2024fast,
  title={Fast contact-implicit model predictive control},
  author={Le Cleac'h, Simon and Howell, Taylor A and Yang, Shuo and Lee, Chi-Yen and Zhang, John and Bishop, Arun and Schwager, Mac and Manchester, Zachary},
  journal={IEEE Transactions on Robotics},
  volume={40},
  pages={1617--1629},
  year={2024},
  publisher={IEEE}
}

@article{shirai2023tactile,
  title={Tactile tool manipulation},
  author={Shirai, Yuki and Jha, Devesh K and Raghunathan, Arvind U and Hong, Dennis},
  journal={arXiv preprint arXiv:2301.06698},
  year={2023}
}

@article{yang2025physics,
  title={Physics-driven data generation for contact-rich manipulation via trajectory optimization},
  author={Yang, Lujie and Suh, HJ and Zhao, Tong and Graesdal, Bernhard Paus and Kelestemur, Tarik and Wang, Jiuguang and Pang, Tao and Tedrake, Russ},
  journal={arXiv preprint arXiv:2502.20382},
  year={2025}
}

@misc{leeIROS2025,
      title={Trajectory Optimization for In-Hand Manipulation with Tactile Force Control}, 
      author={Haegu Lee and Yitaek Kim and Victor Melbye Staven and Christoffer Sloth},
      year={2025},
      eprint={2503.08222},
      archivePrefix={arXiv},
      primaryClass={cs.RO},
      url={https://arxiv.org/abs/2503.08222}, 
}

@inproceedings{xie2016rigid,
  title={Rigid body dynamic simulation with line and surface contact},
  author={Xie, Jiayin and Chakraborty, Nilanjan},
  booktitle={2016 IEEE International Conference on Simulation, Modeling, and Programming for Autonomous Robots (SIMPAR)},
  pages={9--15},
  year={2016},
  organization={IEEE}
}

@inproceedings{dietz2024high,
  title={High Accuracy Numerical Optimal Control for Rigid Bodies with Patch Contacts through Equivalent Contact Points},
  author={Dietz, Christian and Nurkanovi{\'c}, Armin and Albrecht, Sebastian and Diehl, Moritz},
  booktitle={2024 IEEE 63rd Conference on Decision and Control (CDC)},
  pages={901--908},
  year={2024},
  organization={IEEE}
}

@inproceedings{kim2022contact,
  title={Contact-implicit differential dynamic programming for model predictive control with relaxed complementarity constraints},
  author={Kim, Gijeong and Kang, Dongyun and Kim, Joon-Ha and Park, Hae-Won},
  booktitle={2022 IEEE/RSJ International Conference on Intelligent Robots and Systems (IROS)},
  pages={11978--11985},
  year={2022},
  organization={IEEE}
}

@article{oller2024tactile,
  title={Tactile-driven non-prehensile object manipulation via extrinsic contact mode control},
  author={Oller, Miquel and Berenson, Dmitry and Fazeli, Nima},
  journal={arXiv preprint arXiv:2405.18214},
  volume={3},
  year={2024}
}

@inproceedings{hogan2020tactile,
  title={Tactile dexterity: Manipulation primitives with tactile feedback},
  author={Hogan, Francois R and Ballester, Jose and Dong, Siyuan and Rodriguez, Alberto},
  booktitle={2020 IEEE international conference on robotics and automation (ICRA)},
  pages={8863--8869},
  year={2020},
  organization={IEEE}
}

@article{zeng2017energy,
  title={Energy-efficient UAV communication with trajectory optimization},
  author={Zeng, Yong and Zhang, Rui},
  journal={IEEE Transactions on wireless communications},
  volume={16},
  number={6},
  pages={3747--3760},
  year={2017},
  publisher={IEEE}
}

@article{kelly2017introduction,
  title={An introduction to trajectory optimization: How to do your own direct collocation},
  author={Kelly, Matthew},
  journal={SIAM review},
  volume={59},
  number={4},
  pages={849--904},
  year={2017},
  publisher={SIAM}
}
\end{document}